\documentclass[10pt,twocolumn,letterpaper]{article}

\usepackage{cvpr}
\usepackage{times}
\usepackage{epsfig}
\usepackage{graphicx}
\usepackage{amsmath}
\usepackage{amssymb}

\usepackage{subfig}

\usepackage[pagebackref=true,breaklinks=true,letterpaper=true,colorlinks,bookmarks=false]{hyperref}

\cvprfinalcopy % *** Uncomment this line for the final submission

\def\cvprPaperID{583} % *** Enter the CVPR Paper ID here

\ifcvprfinal\fi
\begin{document}

%%%%%%%%% TITLE
\title{Convolutional Regression for Visual Tracking}

\author{Kai Chen, Wenbing Tao\\
School of Automation, Huazhong University of Science and Technology\\
Wuhan, China\\
{\tt\small \{chkap, wenbingtao\}@hust.edu.cn}
% For a paper whose authors are all at the same institution,
% omit the following lines up until the closing ``}''.
% Additional authors and addresses can be added with ``\and'',
% just like the second author.
% To save space, use either the email address or home page, not both
}

\maketitle
%\thispagestyle{empty}

%%%%%%%%% ABSTRACT
\begin{abstract}
   Recently, discriminatively learned correlation filters (DCF) has drawn much attention in visual object tracking community. The success of DCF is potentially attributed to the fact that a large amount of samples are utilized to train the ridge regression model and predict the location of object. To solve the regression problem in an efficient way, these samples are all generated by circularly shifting from a search patch. However, these synthetic samples also induce some negative effects which weaken the robustness of DCF based trackers.
   
   In this paper, we propose a Convolutional Regression framework for visual tracking (CRT). Instead of learning the linear regression model in a closed form, we try to solve the regression problem by optimizing a one-channel-output convolution layer with Gradient Descent (GD). In particular, the receptive field size of the convolution layer is set to the size of object. Contrary to DCF, it is possible to incorporate all ``real" samples clipped from the whole image. A critical issue of the GD approach is that most of the convolutional samples are negative and the contribution of positive samples will be suppressed. To address this problem, we propose a novel Automatic Hard Negative Mining method to eliminate easy negatives and enhance positives. Extensive experiments are conducted on a widely-used benchmark with 100 sequences. The results show that the proposed algorithm achieves outstanding performance and outperforms almost all the existing DCF based algorithms.
%   
%   The regression model can be easily implemented using a one-channel-output convolution layer, of which the receptive field size is set to the size of the object. Then the coefficients of the regression model can be optimized as in a typical C
%
%   
%   The receptive field size of the convolution layer is set to the same of object. Contrary to DCF, the samples here are all "real", i.e.\ they are densely clipped from the image. A critical issue of the GD approach is that most of these "real" samples are negative, and then the contribution of positive samples will be submerged under the negatives. To alleviate this issue, we propose a novel Automatic Hard Negative Mining method by minimizing a truncated loss function. Extensive experiments are conducted on a widely-used benchmark with 100 sequences. The results show that the proposed algorithms achieves outstanding performance compared to other state-of-the-art trackers.
   
\end{abstract}

%%%%%%%%% BODY TEXT
\section{Introduction} \label{sec:intro}

In a generic visual tracking task, the goal is to predict the state of an object, which is labeled in the initial given boundary box, in an image sequence. Compared to a common computer vision problem, visual tracking is very special that the amount of positive samples is quite limited, whereas the negatives are virtually unlimited. So it will always be better for a tracker to make full use of as many negative samples as possible. 

\begin{figure}[!t]
	\centering
	\includegraphics[width=1.0\linewidth]{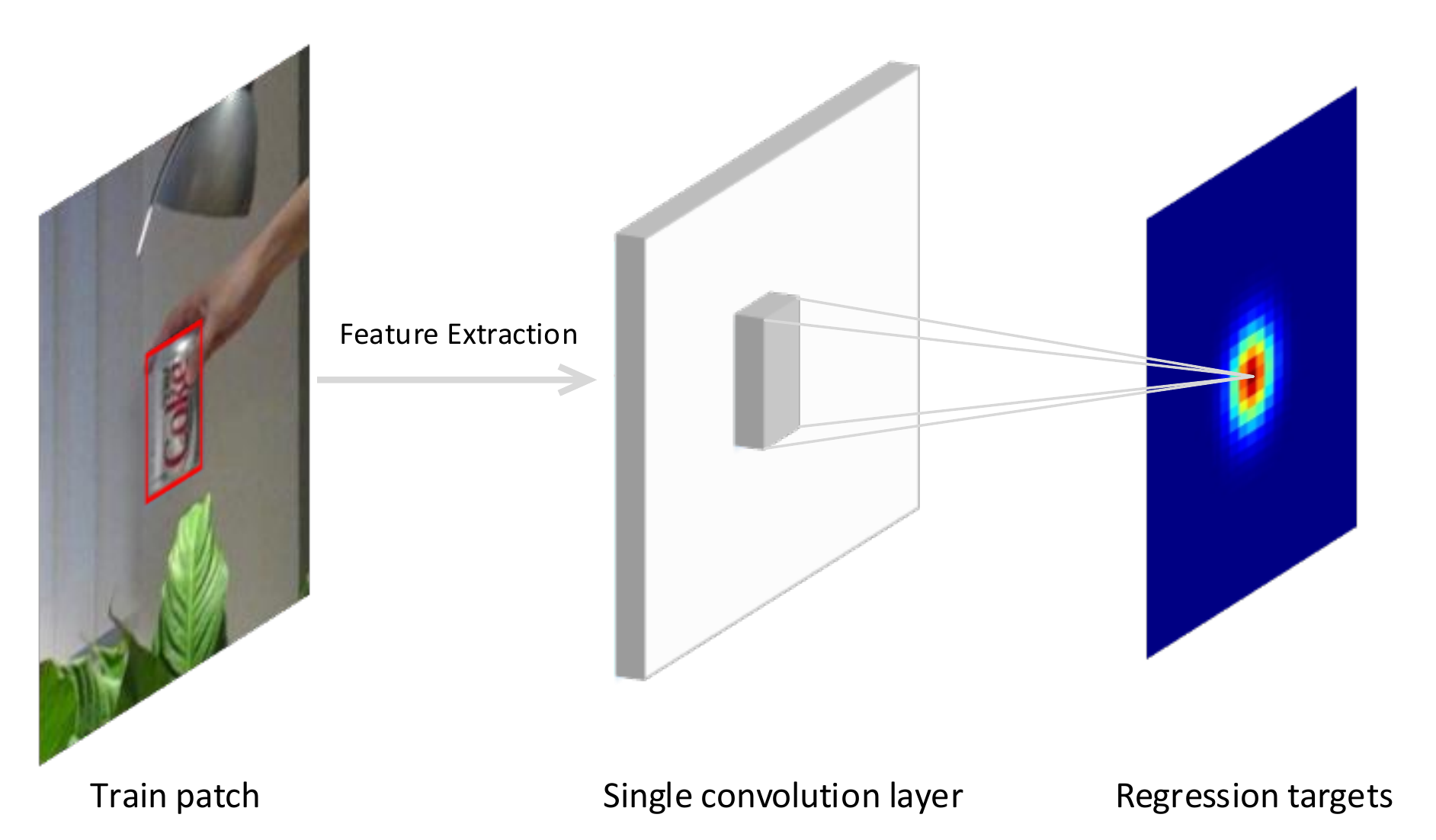}
	\caption{Regression via a single convolution layer. The regression of samples extracted by sliding a window over an image patch can be computed via a single convolution layer. Then the coefficients can be optimized using Gradient Descent together with the back-propagation technique. Compared to the conventional Discriminative Correlation Filters, the Convolutional Regression is trained on ``real" samples with no background context included, and unlimited negative samples can be incorporated.}
	\label{fig:conv_reg}
\end{figure}

In recent years, discriminative algorithms have play an important role in visual tracking. In a discriminative approach, the algorithm can be divided into two parts. One is to represent the object with either handcrafted features, such as original RGB colors, HOG~\cite{2010_PAMI_HOG} and Color Names~\cite{2014_CVPR_CN}, or deeply learned convolutional features from network like VGGNet~\cite{2015_ICLR_VGG} and ResNet~\cite{2015_ResNet}. The other is to learn a discriminative classifier from the initial image. Here, we focus on the later part. A number of classical classification techniques have been utilized in visual tracking, e.g.\ SVM~\cite{2011_ICCV_STRUCK}, MIL~\cite{2011_PAMI_MIL} and AdaBoost~\cite{2006_BMVC_BOOST}. Recently, Discriminatively learned Correlation Filters (DCF)~\cite{2012_ECCV_CSK, 2015_PAMI_KCF} which is developed from the classical ridge regression model, has achieved great success. It is worth noting that these more developed classification techniques like SVM and AdaBoost, which dominate in computer vision tasks like image classification and object detection, however do not perform better than the simpler regression model. It seems that we should focus on somewhere else other than developing an advanced classifier.

The basic idea of DCF is to learn a ridge regression classifier. As we all know, there exists a closed-form analytic solution for the classical regression problem. However, the solution becomes computationally prohibitive when it comes to a large amount of training samples with high feature dimensions. In~\cite{2012_ECCV_CSK, 2015_PAMI_KCF}, a workaround is proposed to generate the samples by shifting from one search patch. In consideration of the inherent correlation between these cyclic samples, the classical solution can be further simplified to reduce the time complexity. The DCF incorporates more than thousands of samples for both training and detecting, which is virtually impossible in a conventional discriminative classifier. It implies that the key point to build a robust tracker is to make full use of as many samples as possible. However, the workaround in DCF also induces some negative effects. (a) The samples for training and detecting are all synthetic, which may decrease the effectiveness of the regression model. (b) Too much background information is included in the samples, which will disturb the prediction for positive samples. (c) The search space is limited to the size of the sample. These three negative effects will significantly limit the performance of DCF. The recent SRDCF~\cite{2015_ICCV_SRDCF} alleviates the issues by introducing a spatial regularization component in the original DCF formulation, and improves the performance with a large margin. However, this does not solve the problem fundamentally.

In this paper, we try to address these issues in a different way by proposing a novel Convolutional Regression framework for visual tracking (CRT), as shown in figure \ref{fig:conv_reg}. Instead of looking for an analytic solution to the regression problem, we try to obtain an approximate solution via Gradient Descent (GD). In our framework, the regression model is built over a one-channel-output convolution layer, as used in a typical Convolutional Neural Networks (CNN) except that the receptive field size is set to the size of object. Then the coefficients, i.e.\ the weight and bias parameters for this convolution layer, can be optimized by minimizing a loss function of the convolution output. The model can be easily implemented in a modern machine learning framework like TensorFlow~\cite{2015_tensorflow} and Caffe~\cite{2014_caffe}. Unlike DCF and SRDCF, it is possible to incorporate virtually all real samples extracted by sliding a window over the whole image during training and detecting stages. A critical issue of CRT is that more than 95 percent of the training samples are negative, and the regression model would be over fitted to predict negatives. To deal with this issue, we propose an Automatic Hard Negative Mining method by introducing a truncated loss function to eliminate easy negatives and a weighted function to enhance positives. The proposed method significantly speed up the training stage, compared to the standard approach.

We perform extensive experiments on two frequently used datasets: OTB-100~\cite{2015_PAMI_OTB} with 100 sequences and OTB-50~\cite{2015_PAMI_OTB} (a more challenging subset of OTB-100). Our proposed algorithm outperforms, as far as we know, all the existing DCF or SRDCF based trackers and other state-of-the-art trackers in both OTB-100 and OTB-50.

\section{Related Work}
In this section, we firstly introduce the Discriminative Correlation filter based trackers which are basically solving a ridge regression problem. Then we introduce the CNN based trackers, since the proposed Convolutional Regression is, in some degree, a special Fully Convolutional Network.

\subparagraph{DCF based trackers}
Discriminative Correlation filters have been fully exploited to build efficient and robust trackers. The core of DCF is that a large amount of samples can be utilized for both training and detecting. Correlation filter is firstly introduced into visual tracking by the MOSSE tracker~\cite{2010_CVPR_MOSSE}, in which only a single-channel feature is adopted. In~\cite{2012_ECCV_CSK}, kernelized correlation filter with circularly generated samples is proposed, and is further improved in~\cite{2015_PAMI_KCF} by adopting the HOG features. A number of trackers~\cite{2015_CVPR_LCT, 2015_CVPR_RPT, 2015_CVPR_MUSTer} developed from the DCF framework have been proposed to improve the performance. However the above mentioned trackers all fail to resolve the issues originating from the circular structure in DCF, as mentioned in section~\ref{sec:intro}. In~\cite{2015_ICCV_SRDCF}, Danelljan et al.\ propose to alleviate these issues by introducing a spatial regularization component to penalize the coefficients of background context in search patch. In the SRDCF framework, the search patch can be much larger than in DCF, so that more negative samples can be utilized. As a result, the SRDCF based tracker achieves a significant improvement compared to the other DCF trackers. In~\cite{2015_ICCVW_DeepSRDCF}, the performance is further improved by adopting deeply learned convolutional features. In this paper, we try to address the issues in a different way.

\subparagraph{CNN based trackers}
Benefiting from large scale training dataset like ImageNet~\cite{2015_IJCV_ILSVRC15}, CNN has achieved great success in computer vision tasks like image classification and object detection. In visual tracking, it is generally impossible to train a deep CNN because of the quite limited training data. Instead, we can transfer a deep CNN like VGGNet~\cite{2015_ICLR_VGG} trained for image classification to extract convolutional features for visual tracking. In~\cite{2015_ICCV_HCF}, both shallow and deep convolutional features extracted from a pre-trained CNN are utilized in the DCF framework. Wang et al.\ \cite{2015_ICCV_GS} propose a two-stream fully convolutional network to capture both general object information and specific discriminative information for visual tracking. In~\cite{2016_CVPR_HDT}, Qi et al.\ propose an adaptive Hedge method to hedge different CNN trackers into a stronger one.

\section{Convolutional Regression} \label{sec:cr}
In this section, we describe the approach to learn a regression model with a single convolution layer in detail.

\subsection{Regression via Convolution Layer}

At first, we have a review of how a linear ridge regression model can be exploited for visual tracking, as in DCF~\cite{2012_ECCV_CSK, 2015_PAMI_KCF}. Given an initial image with labeled target, we can extract numerous training samples $X \in \mathbb{R}^{m\times n}$, and also the corresponding regression targets $Y \in \mathbb{R}^{m}$. Here, $m$ is the number of training samples, and $n$ is the dimension of sample features. Each row of $X$ denotes one sample $x_i$, and the corresponding regression target is $y_i$, the $i$th element of $Y$. Then, the goal is to learn the coefficients $w$ for the regression function $f(z)=w^T\cdot z$, by minimizing the following objective function,

\begin{equation} \label{eq:ridge_regression}
\mathop{\arg\min}_{w}{\lVert X\cdot w - Y \rVert^2 + \lambda\lVert w \rVert^2}.
\end{equation}
Here, $\lVert \cdot \rVert$ means the Euclidean norm, and $\lambda$ is a regularization parameter that controls overfitting. There exists a closed-form analytic solution for this problem,
\begin{equation} \label{eq:rr_solve}
w=(X^TX+\lambda I)^{-1}X^TY.
\end{equation}
However, solving the regression problem with equation (\ref{eq:rr_solve}) becomes computationally prohibitive when $m$ and $n$ is large, e.g.\ more than 1000, which is usually normal for visual tracking. This will absolutely limit the application of ridge regression in visual tracking. In DCF, a workaround is proposed by generating the samples by circularly shifting from a search patch. Then equation (\ref{eq:rr_solve}) can be simplified for efficient computation. Here, we try to solve the regression problem in a different way.

Inspired by the great success of Convolutional Neural Networks (CNN) and the highly developed machine learning frameworks like Tensorflow and Caffe, we propose to learn a regression model with Gradient Descent (GD). Instead of generating cyclic samples from one single patch as in DCF, we extract training and predicting samples by sliding windows over the given image. Then the regression results of these samples can be calculated via a convolution layer with one-channel output. Different from a conventional convolution layer, of which the receptive field size is usually $3\times3$ or $5\times5$ for extracting convolutional features, we set the receptive field size to the size of the object for tracking. The gradients of the coefficients $w$, i.e.\ the weight and bias parameters of this convolution layer, can be calculated by back-propagating the total loss defined in equation (\ref{eq:ridge_regression}), which can be implemented in almost all modern machine learning frameworks. Then an approximate optima of $w$ can be obtained by iteratively applying the Gradient Descent technique.

Compared to DCF, our approach has three advantages. (a) The samples for both training and detecting are all extracted with no background context included. This is helpful to improve performance especially for scale estimation. (b) Virtually unlimited negative samples throughout the whole image can be exploited for training and updating the regression model. This will significantly decrease the probability of drifting from the object even when the object is occluded. (c) The search space for detecting the object is technically unlimited, which is important in case of fast motion.

\subsection{Automatic Hard Negative Mining} \label{sec:AHNM}

Note that the objective function defined in equation (\ref{eq:ridge_regression}) is convex, it is possible to obtain the global optima via Gradient Descent with a small enough learning rate in limited steps. However, it will usually take a long time, which makes it impracticable for visual tracking. Moreover, the positive samples in visual tracking are very limited, while the negative samples are virtually unlimited. As in our following experiments, more than 95 percent of these samples will be negative. As a result, the contributions of positive samples would be submerged under the dominant negatives. It will be difficult to train the regression model to predict the positives accurately.

To address the issues, we propose an novel Automatic Hard Negative Mining method. Considering that most of the negatives can be predicted with low regression error, we propose a truncated loss function to eliminate these easy negatives, defined as
\begin{equation} \label{eq:truncated_loss}
	\text{T}(e)=\left\{ 
		\begin{array}{ll}
			e, & \qquad\text{if } |e|\geqslant th \\
			0, & \qquad\text{otherwise}
		\end{array}
		\right.
\end{equation}
where $th$ is a manually set threshold. By applying this function to the regression errors, i.e.\ the first term in equation (\ref{eq:ridge_regression}), the contributions of these easy negatives to update the coefficients $w$ using GD will be eliminated. The hard negatives and the positives will potentially not be affected.

Furthermore, we introduce a weight function to enlarge the contributions of positives. The motivation of this weight function comes from the fact that it is more important to predict positives accurately than negatives. The weight function is defined as
\begin{equation} \label{eq:weight_func}
\text{W}(y)=\exp(a\cdot y).
\end{equation}
Here, $y$ denotes the regression target of a sample. Usually, a positive sample is labeled with a higher $y$ than a negative sample.

Finally, the improved objective function can be defined as
\begin{equation} \label{eq:improved_regression}
\mathop{\arg\min}_{w}{\lVert \text{W}(Y) \odot \text{T}\left( X\cdot w - Y\right) \rVert^2 + \lambda\lVert w \rVert^2},
\end{equation}
where $\odot$ means the Hadamard product. Note that when $th$ is set to $0$ and $a$ is set to $0$, equation (\ref{eq:improved_regression}) degrades into equation (\ref{eq:ridge_regression}). The truncated function $\text{T}(e)$ and the weight function $\text{W}(y)$ all can be implemented using the built-in operations in TensorFlow.

To validate the proposed method, we perform experiments to evaluate the converging speed of different settings of $th$ and $a$. At first, we need to define a metric for evaluating the regression results after each train step. Considering that it is much more important to predict positives accurately rather than negatives for a regressor in visual tracking, we use the Signal-to-Noise Ratio (SNR) to measure how well the object is predicted. Here, the Signal means the regression result of the object patch, and the Noise means the mean regression results of background patches. Let $M$ denote the convolutional regression results of a given train patch, then the SNR can be defined as,
\begin{equation}
\text{SNR}(M) = \exp(\text{max} (M) - \text{mean}(M))
\end{equation}
For simplicity, the maximum of $M$ is used to approximate the Signal, and the mean of $M$ is to approximate the Noise. 

\begin{figure}[!t]
	\centering
	\includegraphics[width=1.0\linewidth]{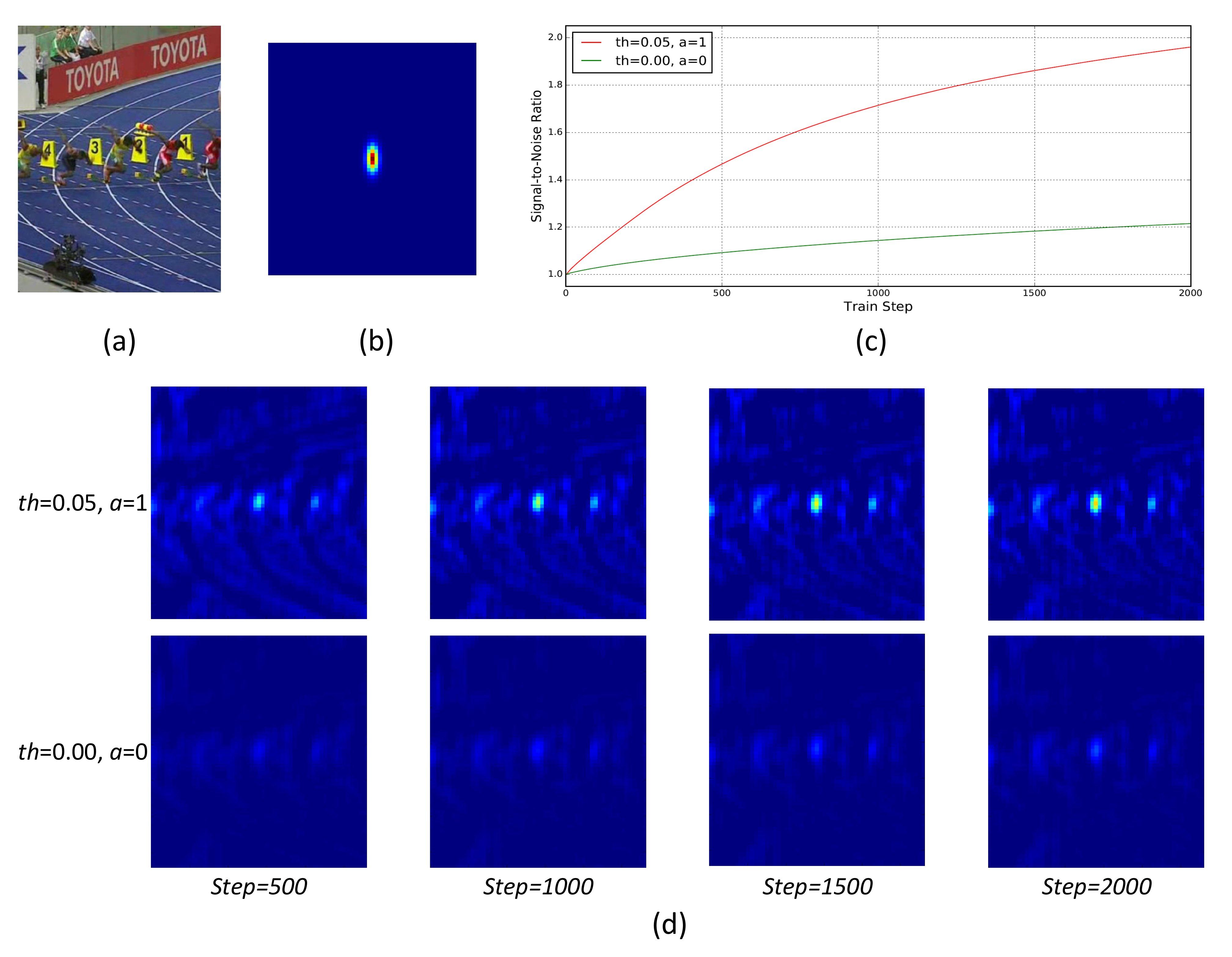}
	\caption{(a) is the train patch used to evaluate the converging speed of Gradient Descent. (b) is the corresponding regression targets. The results of different configurations are plotted in (c). For better visual comparison, the snapshots of regression results are also provided in (d). It can be seen that, the proposed method significantly improves the converging speed, compared to the conventional approach.}
	\label{fig:training}
\end{figure}

The converging speed of three different settings are evaluated on one train patch, as shown in figure \ref{fig:training}. In this experiment, the standard gradient descent optimizer is used to update the coefficients. And the learning rate is fixed to 1e-10, the regularization parameter $\lambda$ is fixed to 1e3. The plots of Signal-to-Noise Ratio versus train step in figure~\ref{fig:training}c shows that, by enabling the truncated function and the weighting function, the converging speed can be significantly improved in terms of the defined SNR. As shown in figure \ref{fig:training}d, the conventional approach with $th=0.00, a=0$ is to reduce the regression errors of all samples including numerous easy negatives. However, the regression for the positive minority is suppressed. By eliminating the easy negatives with $th=0.05$ and enhancing the weights of positives with $a=1$, the positives can be predicted more accurately.

%\begin{figure}[!t]
%	\centering
%	\subfloat[][]{
%		\includegraphics[width=0.18\linewidth]{training/search_bgr}
%		}
%	\subfloat[][]{
%		\includegraphics[width=0.2\linewidth]{training/label}
%		}
%	\subfloat[][]{
%		\includegraphics[width=0.5\linewidth]{training/snr}
%		} \\
%	\subfloat[][]{
%		\includegraphics[width=0.8\linewidth]{training/snapshots}
%		}
%	\caption{(a) is the train patch used to evaluate the converging speed of Gradient Descent. (b) is the corresponding regression targets. The results of different configurations are plotted in (c). For better visual comparison, the snapshots of regression results are also provided in (d). It can be seen that, the proposed method significantly improves the converging speed, compared to the conventional approach.}
%	\label{fig:training}
%\end{figure}

\section{Tracking via Convolutional Regression}
The visual tracking framework based on Convolutional Regression can be decomposed into three stages, i.e.\ training, detecting, updating. We explain each part in three paragraphs, respectively.

\subparagraph{Training} For each sequence, we firstly clip a train patch centered at the given object from the initial image. Since the background in the tracking sequence is usually static, the train patch should be much larger than the object to cover as many background information as possible, so as to decrease the probability of drifting from the object in the following frames. Then a Convolutional Regression network as described in section \ref{sec:cr} can be built. The training features can be extracted from the train patch using HOG~\cite{2010_PAMI_HOG} or any other deep convolutional networks. And the regression target map can be generated using a Gaussian function with variances proportional to the width and height of object. The coefficients are randomly initialized following a zero-mean Gaussian distribution. Then the generated training data are repeatedly passed into the Convolutional Regression network to update the coefficients until reaching a given loss threshold or a limited step.

\begin{figure}[!t]
	\centering
	\subfloat[Examples of detecting procedure]{
		\includegraphics[width=1.0\linewidth]{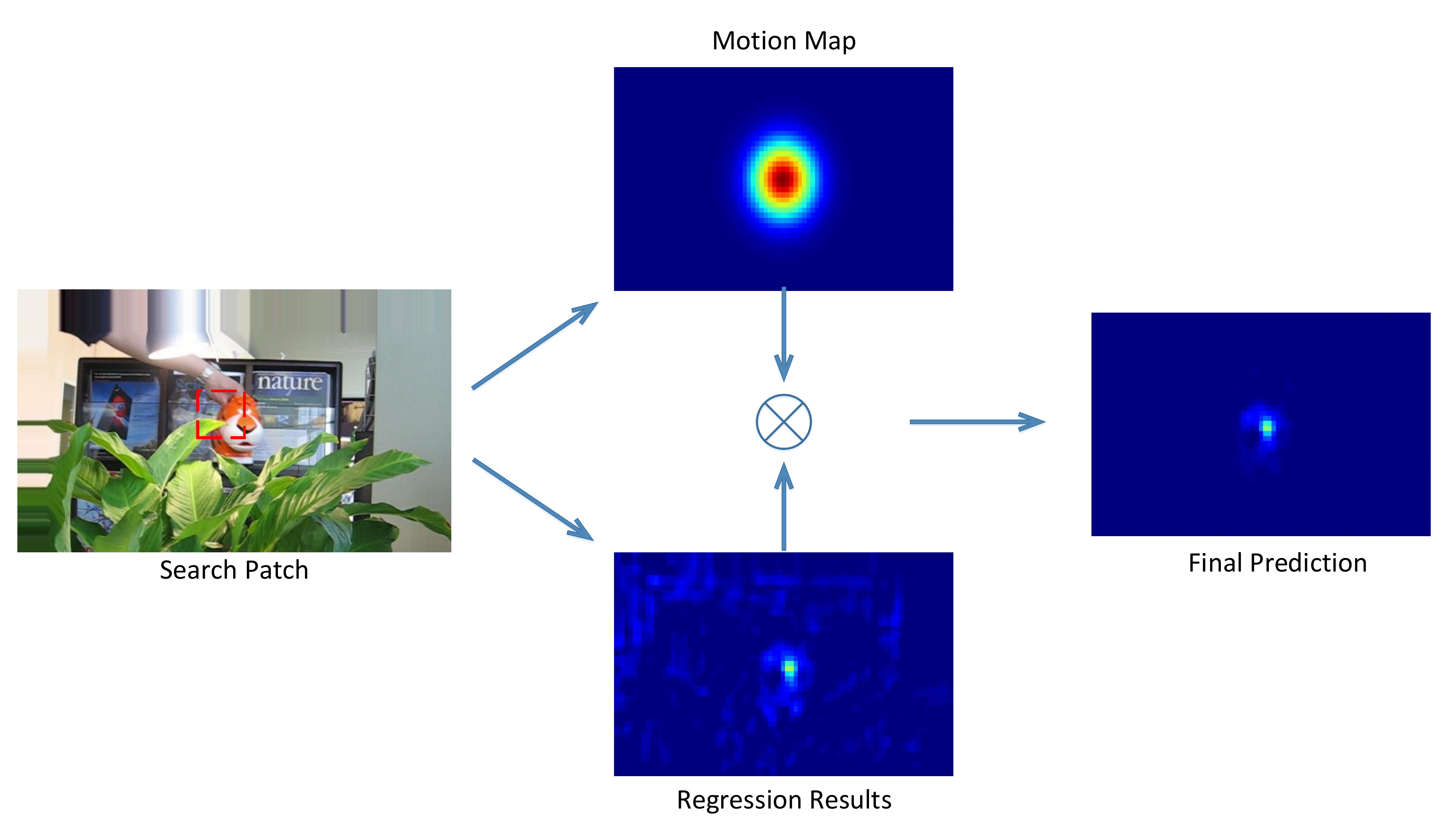}
		\label{fig:detecting}
	} \\
	\subfloat[Examples of updating with historical train data]{
		\includegraphics[width=1.0\linewidth]{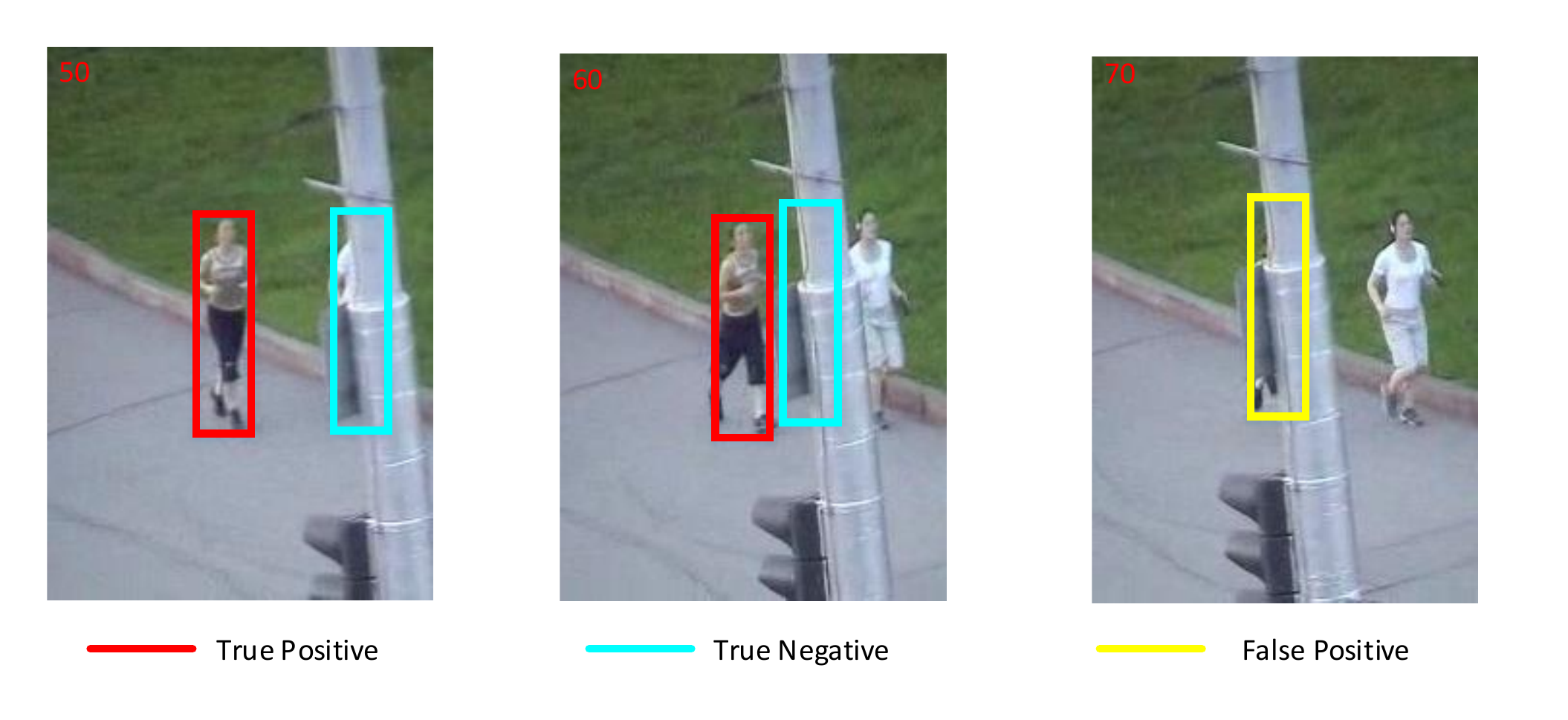}
		\label{fig:updating}
	}
	\caption{(a) The regression results is weighted by a motion map, which is generated according to the location of last object, to eliminate distractors in background. Then the location of object can be predicted by the index of the maximum in the final prediction map. (b) A false positive patch is detected in frame 70, as the person is fully occluded. However, this same patch is labeled as true negative in historical train data, i.e.\ frame 50 and 60. By updating the regression model with historical train data, the impact of false positives can be eliminated.}
\end{figure}

\subparagraph{Detecting} At this stage, a search patch centered at the last object with the same size of the above mentioned train patch is clipped. Then the extracted features of the search patch are passed into the learned Convolutional Regression network to obtain the regression results, i.e.\ the one-channel convolutional output. We further introduce prior motion information to increase the stability. A motion map is generated using a Gaussian function with variation proportional to the size of object. A final prediction map is then calculated by multiplying the motion map and the regression results. The index of the maximum of the final prediction map indicates the final location of object. The detecting procedure is shown in figure~\ref{fig:detecting}. For scale estimation, we simply adopt a naive implementation by repeating the above procedure on scaled objects.

\subparagraph{Updating} To deal with the varied object appearance, it is important to update the initially learned regression model. The new train data pair can be  generated according to the location of new tracked object, as in the Training stage. To update the regression model smoothly, train data pairs generated from several past frames are all used. As shown in figure~\ref{fig:updating},  the inclusion of historical train data is helpful to eliminate the impact of wrongly tracked object. In this stage, the coefficients are updated via Gradient Descent in fixed steps.

\section{Experiments}
To evaluate the proposed Convolutional Regression framework for visual tracking, we perform extensive experiments on two frequently used benchmark: OTB-100~\cite{2015_PAMI_OTB} with 100 sequences and OTB-50~\cite{2015_PAMI_OTB}, which is a more challenging subset of OTB-100 with 50 sequences.

\subsection{Experiment setup}
In our experiments, the height of the train patch is set to 5 times the height of object, while the width of train patch is augmented to 9 times the object's, because the width of object is usually much smaller than height. Technically, the train patch can be larger, but it will take more time to extract features and few improvements can be gained. The size of search patch is the same as that of the train patch, so that the extracted features can be re-used in the Updating stage. The regression target map and the motion map are both generated using a two-dimensional Gaussian function with peak value of 1.0, except that the variances for the regression target map are set to 0.1 times the width and height of object respectively, while the variance of the motion map is 0.6 times the size of object. 

The threshold parameter $th$ defined in equation \ref{eq:truncated_loss} is set to 0.1, and the parameter $a$ that controls the weight function in equation \ref{eq:weight_func} is set to 1.0. The regularization parameter $\lambda$ is set to 1e3. In the training stage, we iteratively apply the Adam optimizer \cite{2014_CoRR_Adam} with a learning rate of 2e-8 to update the coefficients, until the total loss defined in equation \ref{eq:improved_regression} is below a given threshold 0.02, or the maximum allowed train step 4000 is reached. Typically, the training stage accomplishes in only several hundred steps, due to the efficient Adam optimizer and the proposed Automatic Hard Negative Mining method. In the updating stage, we update the coefficients using train data generated in 5 past frames for only 2 iterations with learning rate reduced to 5e-9.

The Convolutional Regression based tracker is implemented with Python and TensorFlow. The whole experiments are conducted on a workstation with a GPU of Tesla K40c.

\subsection{Feature Selection}
In this subsection, we evaluate different features on the Convolutional Regression based tracking framework. Conventional features like original RGB colors, HOG~\cite{2010_PAMI_HOG} and Corlor Names \cite{2014_CVPR_CN} can be easily incorporated in our framework. However, these handcrafted features have been outperformed by the recent convolutional features learned from deep CNN like VGGNet \cite{2015_ICLR_VGG} and ResNet \cite{2015_ResNet}, as in \cite{2015_ICCV_HCF,2015_ICCVW_DeepSRDCF,2016_CVPR_HDT}. In this experiment, we transfer the convolution layers in VGGNet \cite{2015_ICLR_VGG} to extract deep convolutional features from image patch with arbitrary size. The network for feature extraction has a similar architecture to the original VGG-D configuration, except that only the first two max-pool layers are retained. In the new network, even the deepest convolution layer will capture enough spatial information. Then, convolutional features can be extracted as the output of a specific convolution layer. Since the feature channels of deep convolution layers in VGG16 are too large to fit the Convolutonal Regression framework, we use the PCA technique to reduce the feature channels to 64. The PCA can be created with the convolutional features extracted from the initial frame.

We evaluate features extracted from 5 different convolution layers of VGG-D and the HOG features on OTB-50. The results are evaluated in terms of overlap success rate (OS) and distance precision (DP). The OS is calculated as the percentage of frames that the intersection-over-union rate between track result and ground truth is larger than a threshold $T=0.5$. And the DP is the percentage of frames that the center location error is smaller than 20 pixels. More details can be found in \cite{2013_CVPR_OTB}. The score of each tracker with different feature configuration is shown in table \ref{tab:feature_comparison}. All the VGG convolutional features outperforms the handcrafted HOG features. The best performance is obtained using features extracted from the 10-th convolution layer, not from last layer, nor from the first layer as in \cite{2015_ICCVW_DeepSRDCF}. The FPS scores indicate that the bottleneck of running speed lies on the PCA computation and the deep convolutional feature extraction.

\begin{table}[t]
	\centering
	\begin{tabular}{p{4cm} c c c}
		\hline
		& OS (\%) & DP (\%)& FPS \\
		\hline \hline
		CRT - VGG-conv-13 & 68.8 & 76.9 & 0.5 \\
		CRT - VGG-conv-10 & \textbf{75.7} & \textbf{83.5} & 1.3 \\
		CRT - VGG-conv-7 & 70.2 & 78.0 & 2.3 \\
		CRT - VGG-conv-4 & 67.3 & 79.5 & 3.0 \\
		CRT - VGG-conv-2 & 68.9 & 76.4 & 8.2 \\
		CRT - HOG & 68.1 & 75.2 & 9.1 \\
		\hline
	\end{tabular}
	\caption{Evaluations of the Convolutional Regression based Tracker (CRT) with different features on OTB-50. The notation ``conv-x" means the features are extracted from the x-th convolution layer.}
	\label{tab:feature_comparison}
\end{table}

\begin{table}[t]
	\centering
	\begin{tabular}{p{5cm} c c}
		\hline
		& OS (\%) & DP (\%)\\
		\hline \hline
		CRT - $th=0.1, a=1$ & \textbf{68.9} & \textbf{76.4}\\
		CRT - $th=0.0, a=0$ & 58.6 & 68.1\\
		\hline
	\end{tabular}
	\caption{Evaluations of different configurations of $th$ and $a$ on OTB-50.}
	\label{tab:AHNM_comparison}
\end{table}

\begin{table}[t]
	\centering
	\begin{tabular}{p{5cm} c c}
		\hline
		& OS (\%) & DP (\%)\\
		\hline \hline
		CRT - width=9x, height=5x & \textbf{68.9} & \textbf{76.4}\\
		CRT - width=7x, height=5x & 66.2 & 73.2\\
		CRT - width=5x, height=3x & 64.1 & 68.6\\
		CRT - width=3x, height=3x & 61.7 & 66.8\\
		\hline
	\end{tabular}
	\caption{Evaluations of different train patch sizes on OTB-50. The notation ``width=ax, height=bx" means the width and height of train patch is a and b times the size of object, respectively.}
	\label{tab:train_patch_comparison}
\end{table}

\begin{figure}[!t]
	\centering
	\subfloat[Results of OPE on OTB-100]{
		\includegraphics[width=1.0\linewidth]{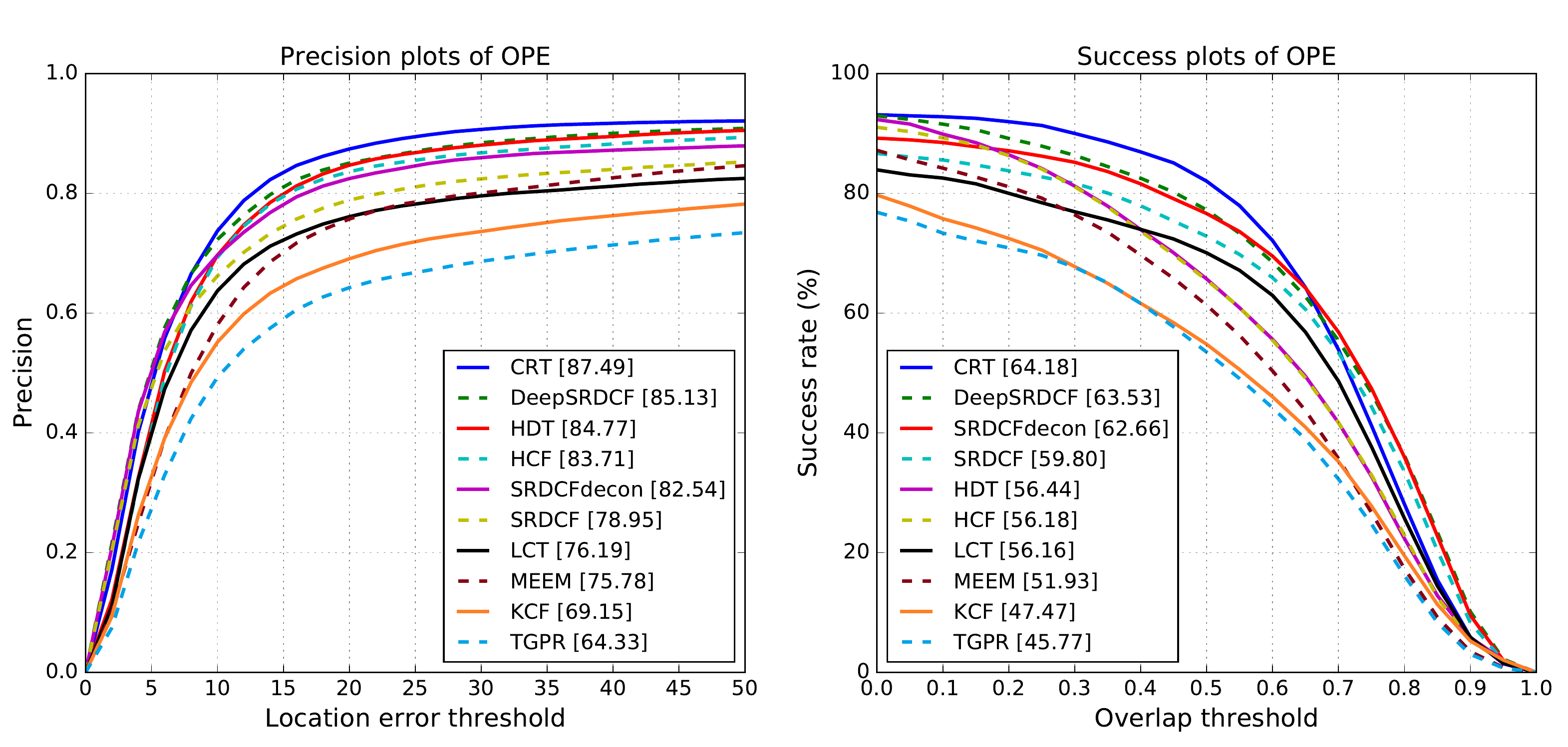}
		\label{fig:otb-100}
	} \\
	\subfloat[Results of OPE on OTB-50]{
		\includegraphics[width=1.0\linewidth]{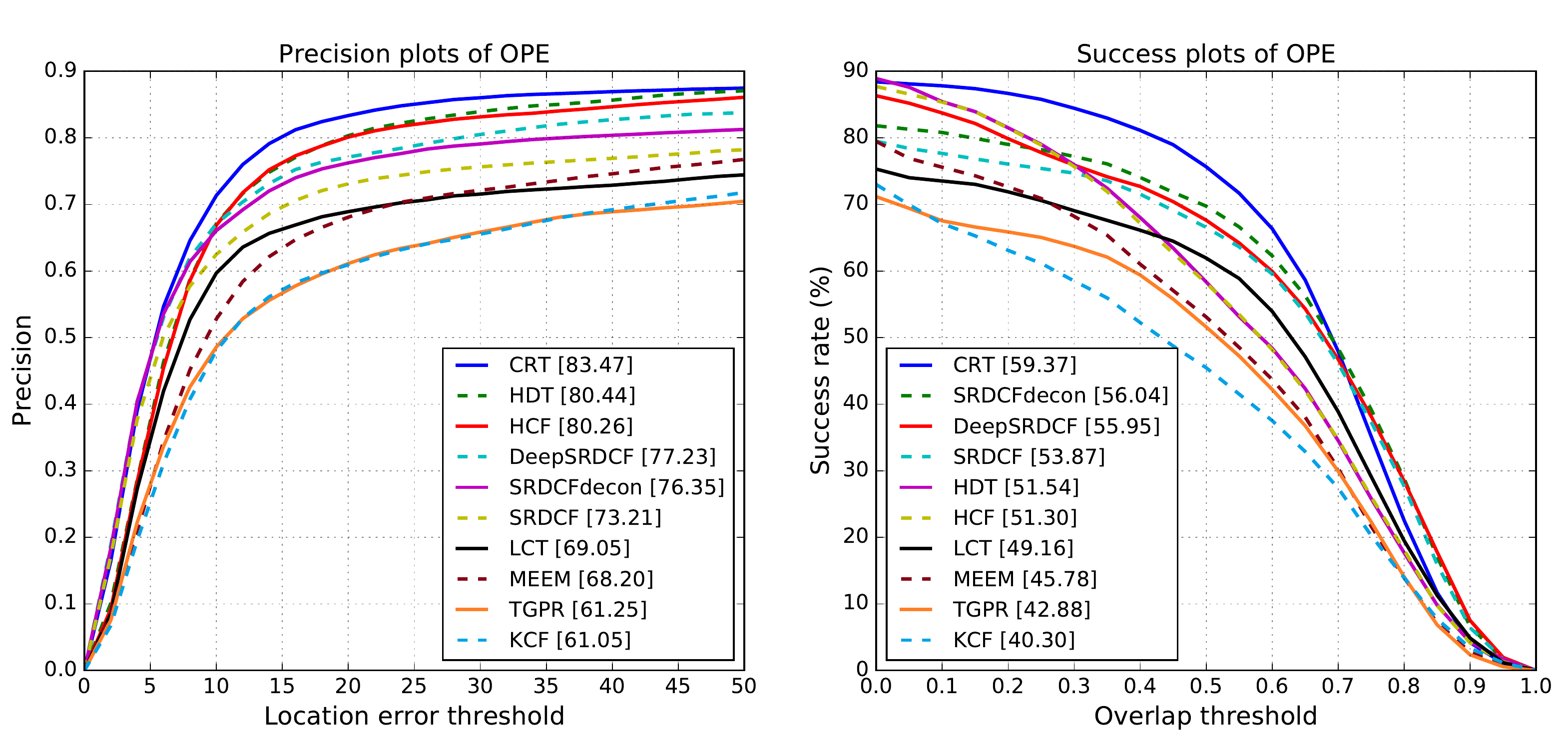}
		\label{fig:otb-50}
	}
	
	\caption{The evaluation results of 10 trackers on OTB-100 and OTB-50. The precision plots are based on the Center Location Error under different distance thresholds, and the numbers in the legends indicate the Distance precision with threshold=20pixel. The success plots are based on the Intersection-Over-Union value under different overlap threshold, and the numbers in the legends indicate the Area-Under-Curve value.}
	\label{fig:result}
\end{figure}

\begin{table*}[!t]
	\centering
	\includegraphics[width=0.95\linewidth]{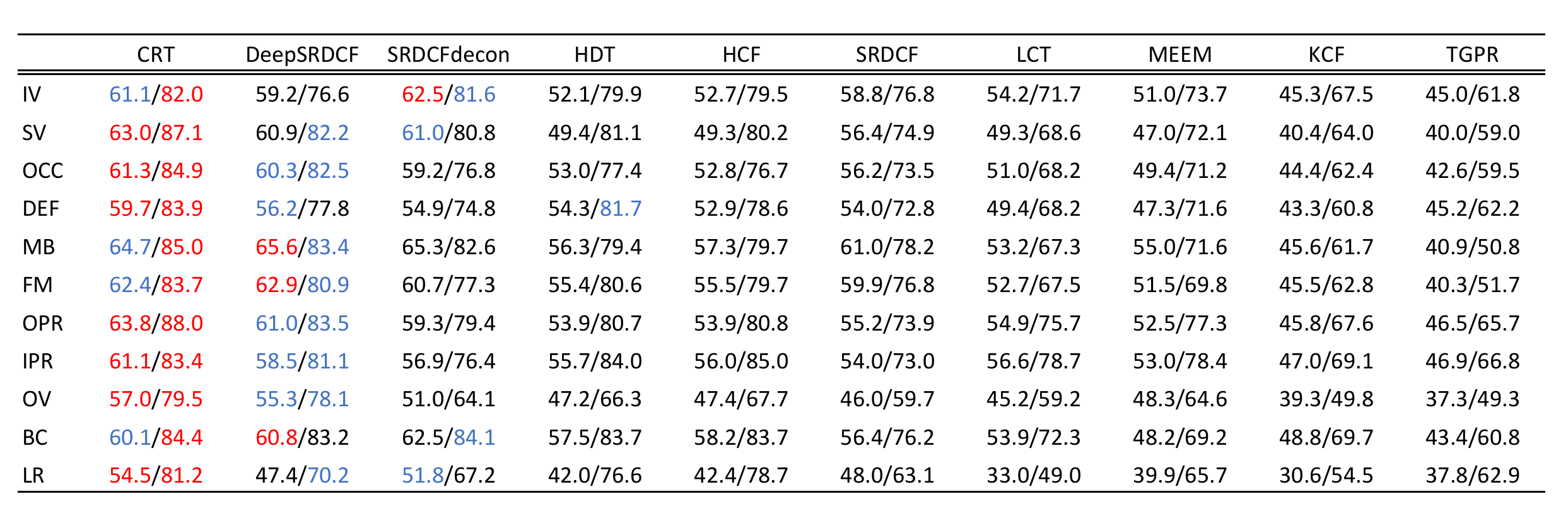}
	\caption{Evaluations under 11 attributes: illumination variation (IV), out-of-plain rotation (OPR), scale variation (SV), occlusion (OCC), deformation (DEF), motion blur (MB), fast motion (FM), in-plain rotation (IPR), out-of-view (OV), background cluttered (BC), and low resolution (LR). The results are shown in the form of ``AUC/DP" where AUC means the Area-Under-Curve value of success plots and DP means the distance precision with threshold=20pixel. The best score is displayed in red color and the second best is in blue color.}
	\label{tab:attrs}
\end{table*}

\subsection{Design Validation}
The key points of our Convolutional Regression based Tracker are that, 1) We propose an Automatic Hard Negative Mining method to eliminate easy negatives and enhance positives. 2) We can incorporate a large amount of negative samples to train the regression model. Here, we perform experiments to validate the two points using features extracted from VGG-conv-2 on OTB-50.

First, we degrade the tracker into the standard approach with objective function defined equation \ref{eq:ridge_regression} by setting $th=0.0, a=0$, and compare it with the proposed CRT tracker (with $th=0.1, a=1$). The results in terms of OS and DP are shown in table \ref{tab:AHNM_comparison}. The proposed Automatic Hard Negative Mining method significantly improves the performance with a gain of 17.6\% in OS and 12.2\% in DP. Additionally, we evaluate the impact of the train patch size. The performances of CRT trackers with different train patch sizes are listed in table \ref{tab:train_patch_comparison}. It is comprehensible that the larger the train patch is, the better the result will be. This just proves that it is important to incorporate as many negative samples as possible. In our approach, the train samples are extracted by sliding a window over the whole train patch, and the widespread negative samples will be conducive to prevent the tracker drifting into background. However, a larger train patch will result in more time consumption for feature extraction and PCA computing. We choose a compromise configuration with ``width=9x and height=5x" for the following experiments.

\subsection{Comparison with state-of-the-art}
We compare the proposed tracker using features extracted from VGG-conv-10 with other 9 state-of-the-art trackers: SRDCFdecon~\cite{2016_CVPR_SRDCFdecon}, DeepSRDCF~\cite{2015_ICCVW_DeepSRDCF}, SRDCF~\cite{2015_ICCV_SRDCF}, HDT~\cite{2016_CVPR_HDT}, HCF~\cite{2015_ICCV_HCF}, LCT~\cite{2015_CVPR_LCT}, MEEM~\cite{2014_ECCV_MEEM}, KCF~\cite{2015_PAMI_KCF}, TGPR~\cite{2014_ECCV_TGPR}. It is worth noting that HDT, HCF, LCT and KCF are all based on the Discriminative Correlation Filters framework, SRDCFdecon, DeepSRDCF and SRDCF are based on the Spatial Regularized Discriminative Correlation Filters framework. In DeepSRDCF, HDT and HCF, deep Convolutional Neural Networks trained on the ImageNet Large Scale Visual Recognition Challenge (ILSVRC) are adopted to extract features for visual tracking.

The experiments are conducted on OTB-100 and OTB-50 using One Pass Evaluation (OPE). The results are evaluated following the standard metrics as proposed in \cite{2015_PAMI_OTB}. As shown in figure \ref{fig:result}, our proposed CRT achieves the first rank in both OTB-100 and OTB-50. The DeepSRDCF is the best Correlation Filter based tracker that incorporates both deep convolutional features and the SRDCF framework. Our CRT outperforms the DeepSRDCF in all the four evaluations. In particular, the performance on OTB-50 is improved by 8\% in the precision plots and 6\% in the success plots. The results have proved that the proposed Convolutional Regression framework is a more effective approach for visual tracking, compared to the SRDCF and DCF framework.

\subsection{Attributes based Comparison}
The main challenges of visual tracking are usually from several aspects. It will be interesting to investigate how well a tracker deals with the variant challenges. In OTB-100, the 100 sequences are all labeled with 11 attributes: illumination variation, out-of-plain rotation, scale variation, occlusion, deformation, motion blur, fast motion, in-plain rotation, out-of-view, background cluttered, and low resolution. We further evaluate the 10 trackers under 11 different attributes.

The evaluation results in terms of AUC and DP are shown in table \ref{tab:attrs}. Among the existing Correlation Filter based trackers, the DeepSRDCF performs well in the tests for Motion Blur (MB), Fast Motion (FM) and Background Cluttered (BC). However, our proposed CRT achieves the first rank in all the 11 attributes in terms of DP. Especially, in Scale Variation (SV), Out-of-Plain Rotation (OPR) and Low Resolution (LR), our CRT outperforms the second best with more than 5 percentage points. The high DP score in all the 11 attributes implies that the CRT can track the object more accurately without drifting into background. This is potentially attributed to the widespread negative samples incorporated in our Convolutional Regression framework.

\begin{figure*}[!t]
	\centering
	\includegraphics[width=0.85\linewidth]{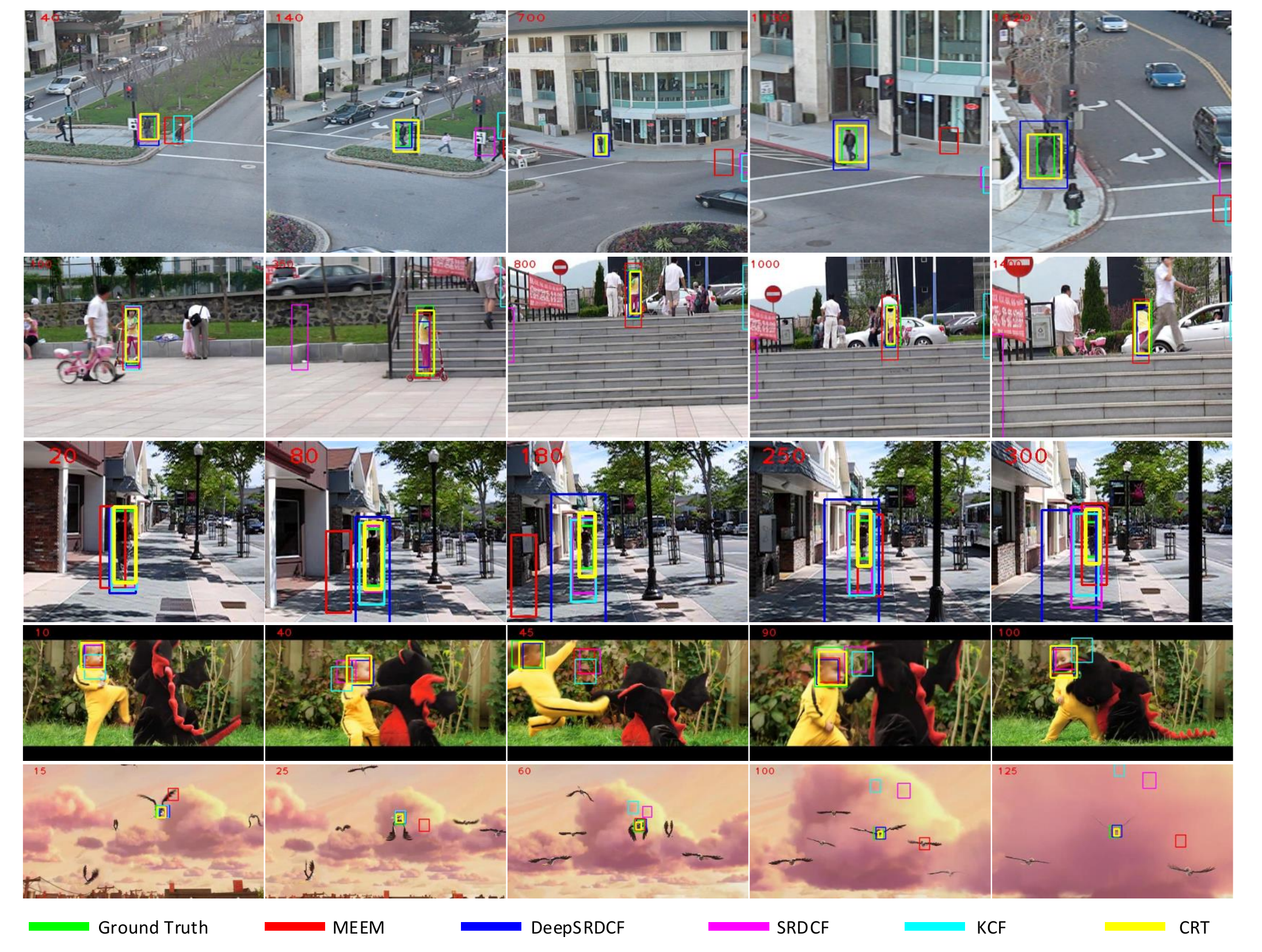}
	
	\caption{Track results of MEEM, DeepSRDCF, SRDCF, KCF and CRT on 5 sequences (\textit{Human3}, \textit{Girl2}, \textit{Human9}, \textit{DragonBaby} and \textit{Bird1}).}
	\label{fig:bbox}
\end{figure*}

\subsection{Qualitative Analysis}

The track results of DeepSRDCF, MEEM, SRDCF, KCF, and the proposed CRT are drew in figure \ref{fig:bbox} for qualitative analysis. In sequences \textit{Human3}, \textit{Girl2} and \textit{Human9}, the targets are rigid, however the background is cluttered. The SRDCF, MEEM and KCF all drift away from the object. The DeepSRDCF successfully tracks the objects, however fails to handle the scale variation in sequences \textit{Human3} and \textit{Human9}. Benefiting from the representation power of deep convolutional feautures, our CRT is able to discriminate the object from the cluttered background. In sequences \textit{DragonBaby} and \textit{Bird1}, the objects undergo severe deformation and rotation. It will be difficult for a classifier to identify the objects correctly. The MEEM, DeepSRDCF, SRDCF and KCF, to some extent, all lost the objects. In this situation, the best solution is to make full use of the known negative samples, so that the tracker will not drift into background. In our CRT, the widespread negative samples around the object are incorporated to update the discriminative model. As a result, the CRT is able to track the objects successfully in \textit{DragonBaby} and \textit{Bird1}.

\section{Conclusion}

In this paper, we propose a novel Convolutional Regression framework for visual tracking. In our algorithm, a linear ridge regression model for visual tracking is trained using Gradient Descent technique by back-propagating regression errors through a single convolution layer. Compared to DCF and SRDCF, our tracking framework can incorporate virtually unlimited ``real" samples. To speed up the training stage, we also propose an Automatic Hard Negative Mining method to eliminate easy negatives and enhance positives. Our extensive experiments show that the proposed method is more effective than DCF and even SRDCF.

{\small
\bibliographystyle{ieee}
\bibliography{crt}
}

\end{document}